\documentclass[conference]{IEEEtran}
\IEEEoverridecommandlockouts

\usepackage[pdftex]{graphicx}
\graphicspath{{./figures/}}

\begin{document}
\title{Neural Decomposition of Time-Series Data for Effective Generalization}

\author{
\IEEEauthorblockN{Luke B. Godfrey and Michael S. Gashler}
\IEEEauthorblockA{Department of Computer Science and Computer Engineering\\
University of Arkansas\\
Fayetteville, AR 72701\\
Email: \{lbg002, mgashler\}@uark.edu}
}

\maketitle

\begin{abstract}
We present a neural network technique for the analysis and extrapolation of time-series data called Neural Decomposition (ND).
Units with a sinusoidal activation function are used to perform a Fourier-like decomposition of training samples into a sum of sinusoids, augmented by units with nonperiodic activation functions to capture linear trends and other nonperiodic components.
We show how careful weight initialization can be combined with regularization to form a simple model that generalizes well.
Our method generalizes effectively on the Mackey-Glass series, a dataset of unemployment rates as reported by the U.S. Department of Labor Statistics, a time-series of monthly international airline passengers, the monthly ozone concentration in downtown Los Angeles, and an unevenly sampled time-series of oxygen isotope measurements from a cave in north India.
We find that ND outperforms popular time-series forecasting techniques including LSTM, echo state networks, ARIMA, SARIMA, SVR with a radial basis function, and Gashler and Ashmore's model.
\end{abstract}

\section{Introduction}

The analysis and forecasting of time-series is a challenging problem that continues to be an active area of research.
Predictive techniques have been presented for an array of problems, including weather \cite{gashler:fourier}, traffic flow \cite{lippi:forecasting}, seizures \cite{elger:seizure}, sales \cite{choi:sarima}, and others \cite{tay:financial,kim:financial,chen:enrollments,taylor:wind}.
Because research in this area can be so widely applied, there is great interest in discovering more accurate methods for time-series forecasting.

One approach for analyzing time-series data is to interpret it as a signal and apply the Fourier transform to decompose the data into a sum of sinusoids \cite{bloomfield2004fourier}.
Unfortunately, despite the well-established utility of the Fourier transform, it cannot be applied directly to time-series forecasting.
The Fourier transform uses a predetermined set of sinusoid frequencies rather than learning the frequencies that are actually expressed in the training data.
Although the signal produced by the Fourier transform perfectly reproduces the training samples, it also predicts that the same pattern of samples will repeat indefinitely.
As a result, the Fourier transform is effective at interpolation but is unable to extrapolate future values.
Another limitation of the Fourier transform is that it only uses periodic components, and thus cannot accurately model the nonperiodic aspects of a signal, such as a linear trend or nonlinear abnormality.

Another approach is regression and extrapolation using a model such as a neural network.
Regular feedforward neural networks with standard sigmoidal activation functions do not tend to perform well at this task because they cannot account for periodic components in the training data.
Fourier neural networks have been proposed, in which feedforward neural networks are given sinusoidal activation functions and are initialized to compute the Fourier transform.
Unfortunately, these models have proven to be difficult to train \cite{gashler:fourier}.

Recurrent neural networks, as opposed to feedforward neural networks, have been successfully applied to time-series prediction \cite{gers2001applying, gers2000learning}.
However, these kinds of networks make up a different class of forecasting techniques.
Recurrent neural networks also have difficulty handling unevenly sampled time-series.
Further discussion about recurrent neural networks and other classes of forecasting techniques is provided in Section~\ref{sec:related}.

We claim that effective generalization can be achieved by regression and extrapolation using a model with two essential properties: (1) it must combine both periodic and nonperiodic components, and (2) it must be able to tune its components as well as the weights used to combine them.
We present a neural network technique called Neural Decomposition (ND) that demonstrates this claim.
Like the Fourier transform, it decomposes a signal into a sum of constituent parts.
Unlike the Fourier transform, however, ND is able to reconstruct a signal that is useful for extrapolating beyond the training samples.
ND trains the components into which it decomposes the signal represented by training samples.
This enables it to find a simpler set of constituent signals.
In contrast to the fast Fourier transform, ND does not require the number of samples to be a power of two, nor does it require that samples be measured at regular intervals.
Additionally, ND facilitates the inclusion of nonperiodic components, such as linear or sigmoidal components, to account for trends and nonlinear irregularities in a signal.

In Section~\ref{sec:results}, we demonstrate that the simple innovations of ND work together to produce significantly improved generalizing accuracy with several problems.
We tested with the chaotic Mackey-Glass series, a dataset of unemployment rates as reported by the U.S. Department of Labor Statistics, a time-series of monthly international airline passengers, the monthly ozone concentration in downtown Los Angeles, and an unevenly sampled time-series of oxygen isotope measurements from a cave in north India.
We compared against long short-term memory networks (LSTM), echo state networks, autoregressive integrated moving average (ARIMA) models, seasonal ARIMA (SARIMA) models, support vector regression with a radial basis function (SVR), and a model recently proposed by Gashler and Ashmore \cite{gashler:fourier}.
In all but one case, ND made better predictions than each of the other prediction techniques evaluated; in the excepted case, LSTM and echo state networks performed slightly better than ND.

This paper is outlined as follows.
Section~\ref{sec:related} provides a background and reviews related works.
Section~\ref{sec:high-level} gives an intuitive-level overview of ND.
Section~\ref{sec:details} provides finer implementation-level details.
Section~\ref{sec:results} shows results that validate our work.
Finally, Section~\ref{sec:conclusion} discusses the contributions of this paper and future work.

\section{Related Work}
\label{sec:related}

\subsection{Models for Time-Series Prediction}

\begin{figure}[!t]
	\centering
	\includegraphics[width=3.5in]{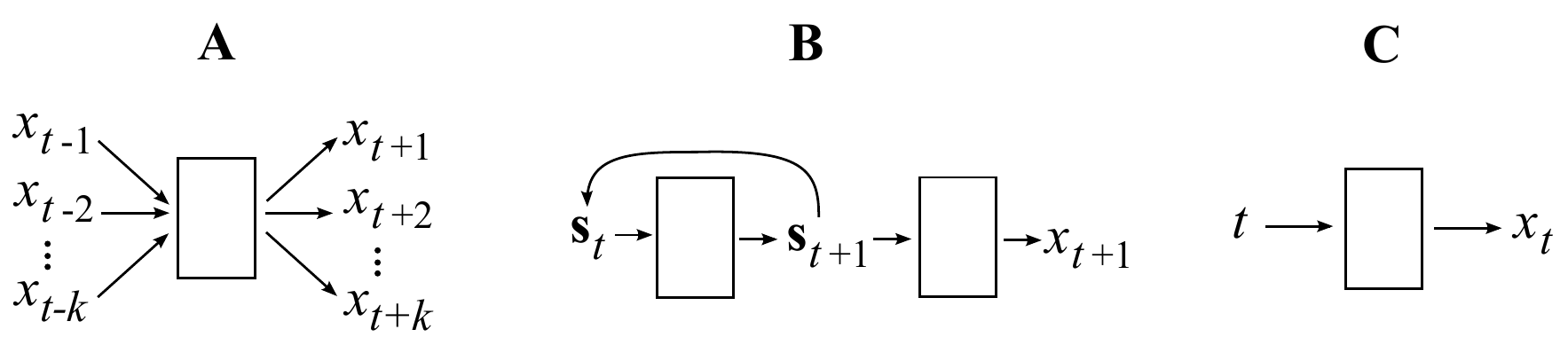}
	\caption{Three broad classes of models for time-series forecasting: (A) prediction using a sliding window, (B) recurrent models, and (C) regression-based extrapolation.}
	\label{fig:types}
\end{figure}

Many works have diligently surveyed the existing literature regarding techniques for forecasting time-series data \cite{dorffner1996neural,kaastra1996designing,zhang1998forecasting,chatfield2000time,frank2001time,zhang2003time,de200625}.
Some popular statistical models include Gaussian process \cite{brahim2004gaussian} and hidden Markov models \cite{macdonald1997hidden}.

Autoregressive integrated moving average (ARIMA) models \cite{zhang:arima,wei1994time} are among the most popular approaches.
The notation for this model is ARIMA($p, d, q$), where $p$ is the number of terms in the autoregressive model, $d$ is the number of differences required to take to make the time-series stationary, and $q$ is the number of terms in the moving average model.
In other words, ARIMA models compute the $d$th difference of $x(t)$ as a function of $x_{t-1}, x_{t-2} , ..., x_{t-p}$ and the previous $q$ error terms.

Out of all the ARIMA variations that have been proposed, seasonal ARIMA (SARIMA) \cite{box:time-series} is considered to be the state of the art ``classical'' time-series approach \cite{lippi:forecasting}.
Notation for SARIMA is ARIMA($p, d, q$)($P, D, Q$)[$S$], where $p, d, q$ are identical to the normal ARIMA model, $P, D, Q$ are analogous seasonal values, and $S$ is the seasonal parameter. For example, an ARIMA(1,0,1)(0,1,1)[12] uses an autoregressive model with one term, a moving average model with one term, one seasonal difference (that is, $x'_t = x_t - x_{t - 12}$), and a seasonal moving average with one term.
This seasonal variation of ARIMA exploits seasonality in data by correlating $x_t$ not only with recent observations like $x_{t-1}$, but also with seasonally recent observations like $x_{t - S}$.
For example, when the data is a monthly time-series, $S=12$ correlates observations made in the same month of different years, and when the data is a daily time-series, $S=7$ correlates observations made on the same day of different weeks.

In the field of machine learning, three high-level classes of techniques (illustrated in Figure~\ref{fig:types}) are commonly used to forecast time-series data \cite{gashler:fourier}.
Perhaps the most common approach, (A), is to train a model to directly forecast future samples based on a sliding window of recently collected samples \cite{frank2001time}.
This approach is popular because it is simple to implement and can work with arbitrary supervised learning techniques.

A more sophisticated approach, (B), is to train a recurrent neural network \cite{jun:traffic,Nerrand94trainingrecurrent}.
Several recurrent models, such as LSTM networks \cite{gers2001applying,gers2000learning}, have reported very good results for forecasting time-series.
In an LSTM network, each neuron in the hidden layer has a memory cell protected by a set of gates that control the flow of formation through time\cite{gers2000learning}.
Echo state networks (ESNs) have also performed particularly well at this task \cite{li2012chaotic,jaeger2004harnessing,shi2007support}.
An ESN is a randomly connected, recurrent reservoir network with three primary meta-parameters: input scaling, spectral radius, and leaking rate \cite{lukosevicius2012esn}.
Although they are powerful, these recurrent models are only able to handle time-series that are sampled at a fixed interval, and thus cannot be directly applied to unevenly sampled time-series.

Our model falls into the third category of machine learning techniques, (C): regression-based extrapolation.
Models of this type fit a curve to the training data, then use the trained curve to anticipate future samples.
One advantage of this approach over recurrent neural networks is that it can make continuous predictions, instead of predicting only at regular intervals, and can therefore be directly applied to irregularly spaced time-series.
A popular method in this category is support vector regression (SVR) \cite{su:traffic,drucker:svr}.
Many models in this category decompose a signal into constituent parts, providing a useful mechanism for analyzing the signal.
Our model is more closely related to a subclass of methods in this category, called Fourier neural networks (see Section~\ref{subsec:fourier_nn}), due to its use of sinusoidal activation functions.
Models in the first two categories, (A) and (B), have already been well-studied, whereas extrapolation with sinusoidal neural networks remains a relatively unexplored area.

\subsection{Harmonic Analysis}

The harmonic analysis of a signal transforms a set of samples from the time domain to the frequency domain.
This is useful in time-series prediction because the resulting frequencies can be used to reconstruct the original signal (interpolation) and to forecast values beyond the sampled time window (extrapolation).
Harmonic analysis, also known as spectral analysis or spectral density estimation, has been well-studied for decades \cite{stein1993harmonic,quinn1989estimating,rubia1999hypofrontality,phillips1999differential}.

Perhaps the most popular method of harmonic analysis is the distrete Fourier transform (DFT).
The DFT maps a series of $N$ complex numbers in the time domain to the frequency domain. The inverse DFT (iDFT) can be applied these new values to map them back to the time domain.
More interestingly, the iDFT can be used as a continuous representation of the originally discrete input.
The transforms are generally written as a sum of $N$ complex exponentials, which can be rewritten in terms of sines and cosines by Euler's formula.

The DFT and the iDFT are effectively the same transform with two key differences.
First, in terms of sinusoids, the DFT uses negative multiples of $2 \pi / N$ as frequencies and the iDFT uses positive multiples of $2 \pi / N$ as frequencies.
Second, the iDFT contains the normalization term $1/N$ applied to each sum.

In general, the iDFT requires all $N$ complex values from the frequency domain to reconstruct the input series.
For real-valued input, however, only the first $N/2 + 1$ complex values are necessary ($N/2$ frequencies and one bias).
The remaining complex numbers are the conjugates of the first half of the values, so they only contain redundant information.
Furthermore, in the real-valued case, the imaginary component of the iDFT output can be discarded to simplify the equation, as we do in Equation~\ref{eq:idft}.
This particular form of the iDFT (reconstructing a series of real samples) can therefore be written as a real sum of sines and cosines.

The iDFT is as follows.
Let $R_k$ and $I_k$ represent the real and imaginary components respectively of the $k$th complex number returned by the DFT.
Let $2 \pi k / N$ be the frequency of the $k$th term.
The first frequency yields the bias, because $cos(0) = 1$ and $sin(0) = 0$.
The second frequency is a single wave, the third frequency is two waves, the fourth frequency is three waves, and so on.
The cosine with the $k$th frequency is scaled by $R_k$, and the sine with the $k$th frequency is scaled by $I_k$.
Thus, the iDFT is sufficiently described as a sum of $N/2 + 1$ terms, with a $sin(t)$ and a $cos(t)$ in each term and a complex number from the DFT corresponding to each term:

\begin{equation}
	x(t) = \sum_{k = 0}^{N/2}{ R_k \cdot cos( \frac{2 \pi k}{N} t ) - I_k \cdot sin( \frac{2 \pi k}{N} t ) }
	\label{eq:idft}
\end{equation}

Equation~\ref{eq:idft} is useful as a continuous representation of the real-valued discrete input.
Because it perfectly passes through the input samples, one might naively expect this function to be a good basis for generalization.
In order to choose appropriate frequencies, however, the iDFT assumes that the underlying function always has a period equal to the size of the samples that represent it, that is, $x(t + N) = x(t)$ for all $t$.
Typically, in cases where generalization is desirable, the period of the underlying function is not known.
The iDFT cannot effectively model the nonperiodic components of a signal, nor can it form a simple model for series that are not periodic at $N$, even if the series is perfectly periodic.

\begin{figure}[!t]
	\centering
	\includegraphics[width=3.5in]{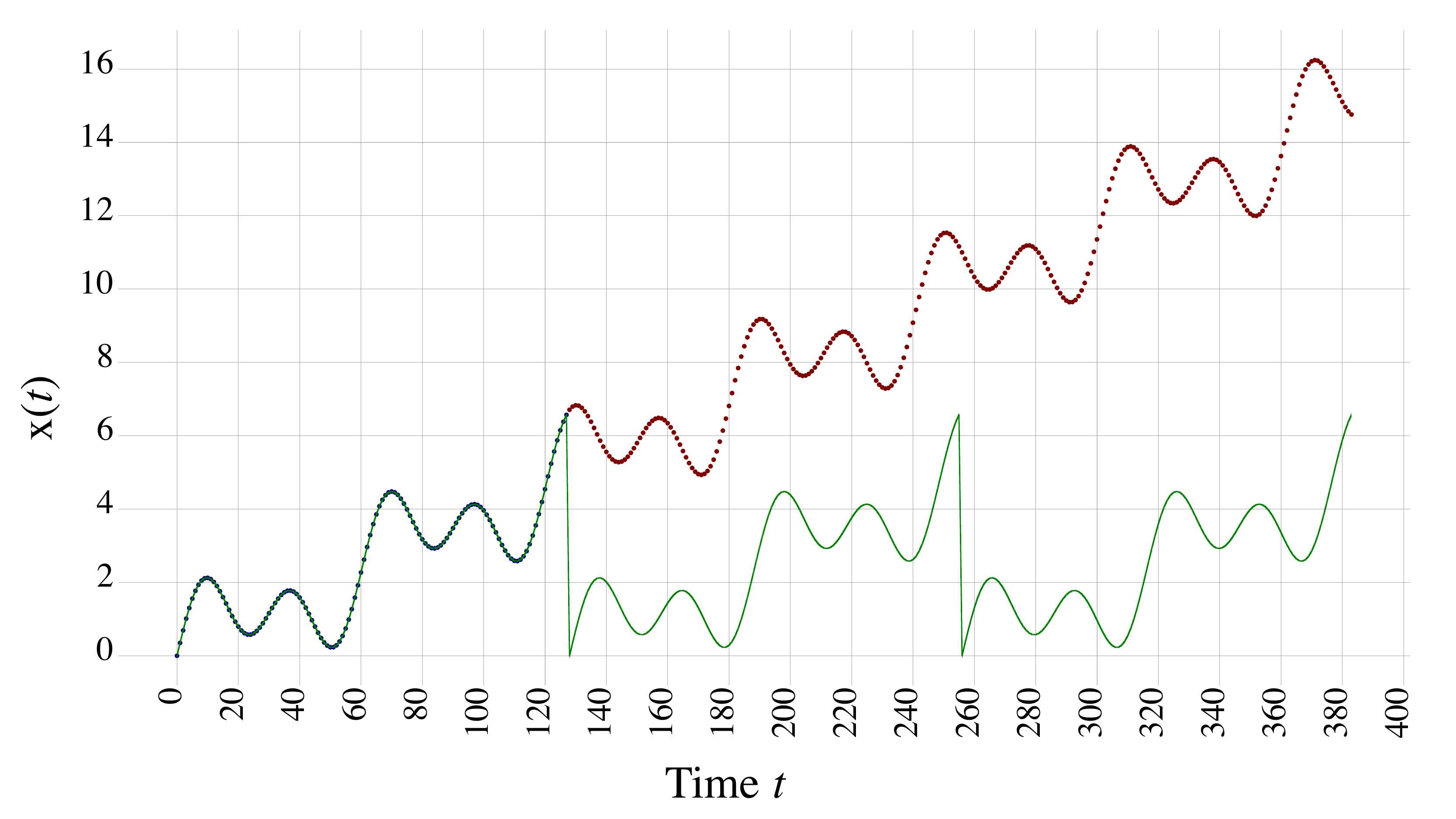}
	\caption{The predictive model generated by the iDFT for a toy problem with both periodic and nonperiodic components. Blue dots represent training samples, red dots represent testing samples, and the green line represents the iDFT. Two significant problems limit its ability to generalize: (1) The model repeats, ignoring the linear trend, and (2) The extrapolated predictions misalign with the phase of the continuing nonlinear trend. }
	\label{fig:idft}
\end{figure}

Figure~\ref{fig:idft} illustrates the problems encountered when using the iDFT for time-series forecasting.
Although the model generated by the iDFT perfectly fits the training samples, it only has periodic components and so is only able to predict that these samples will repeat to infinity, without taking nonperiodicity into account.
Our approach mimics the iDFT for modeling periodic data, but is also able to account for nonperiodic components in a signal (Figure~\ref{fig:toy}).

Because of these limitations of the DFT, other approaches to the harmonic analysis of time-series have been proposed.
Some of these other approaches perform sinusoidal regression to determine frequencies that better represent the periodicity of the sampled signal \cite{vanicek1969petr,yokouchi2000strong}.
Our approach similarly uses regression to find better frequencies.

\subsection{Fourier Neural Networks}
\label{subsec:fourier_nn}

Use of the Fourier transform in neural networks has already been explored in various contexts \cite{hecht1989theory,zuo2009fourier}.
The term \emph{Fourier neural network} has been used to refer to neural networks that use a Fourier-like neuron \cite{silvescu:fourier}, that use the Fourier transform of some data as input \cite{minami:real-time}, or that use the Fourier transform of some data as weights \cite{gashler:fourier}.
Our work is not technically a Fourier neural network, but of these three types, our approach most closely resembles the third.

Silvescu provided a model for a Fourier-like activation function for neurons in neural networks \cite{silvescu:fourier}.
His model utilizes every unit to form DFT-like output for its inputs.
He notes that by using gradient descent to train sinusoid frequencies, the network is able to learn ``exact frequency information'' as opposed to the ``statistical information'' provided by the DFT.
Our approach also trains the frequencies of neurons with a sinusoidal activation function.

Gashler and Ashmore presented a technique that used the fast Fourier transform (FFT) to approximate the DFT, then used the obtained values to initialize the sinusoid weights of a neural network that mixed sinusoidal, linear, and softplus activation functions \cite{gashler:fourier}.
Because this initialization used sinusoid units to model nonperiodic components of the data, their model was designed to heavily regularize sinusoid weights so that as the network was trained, it gave preference to weights associated with nonperiodic units and shifted the weights from the sinusoid units to the linear and softplus units.
Use of the FFT required their input size to be a power of two, and their trained models were slightly out of phase with their validation data.
However, they were able to generalize well for certain problems.
Our approach is similar, except that we do not use the Fourier transform to initialize any weights (further discussion on why we do not use the Fourier transform can be found in Section~\ref{subsec:toy_analysis}).

\section{High Level Approach}
\label{sec:high-level}

In this section, we describe Neural Decomposition (ND), a neural network technique for the analysis and extrapolation of time-series data. This section focuses on an intuitive-level overview of our method; implementation details can be found in Section~\ref{sec:details}.

\subsection{Algorithm Description}

We use an iDFT-like model with two simple but important innovations.
First, we allow sinusoid frequencies to be trained.
Second, we augment the sinusoids with a nonperiodic function to model nonperiodic components.
The iDFT-like use of sinusoids allows our model to fit to periodic data, the ability to train the frequencies allows our model to learn the true period of a signal, and the augmentation function enables our model to forecast time-series that are made up of both periodic and nonperiodic components.

Our model is defined as follows. Let each $a_k$ represent an amplitude, each $w_k$ represent a frequency, and each $\phi_k$ represent a phase shift. Let $g(t)$ be an augmentation function that represents the nonperiodic components of the signal.

\begin{equation}
	x(t) = \sum_{k = 1}^{N}{ \big( a_k \cdot sin( w_k t + \phi_k ) \big) } + g(t)
	\label{eq:gift}
\end{equation}

Note that in our model, compared to the iDFT, two indexing changes have been made:
1) the lower index of the sum has changed from $k = 0$ to $k = 1$, and
2) the upper index of the sum has changed from $N/2$ to $N$.
The lower index has changed because ND can account for bias in the augmentation function $g(t)$, so the 0 frequency is not necessary.
The upper index has changed to simplify the equation as a sum of $N$ sines rather than a sum of $N/2$ sines and cosines.

If the phase shifts are set so that $sin(t + \phi)$ is transformed into $cos(t)$ and $-sin(t)$, the frequencies are set to the appropriate multiples of $2 \pi$, the amplitudes are set to the output values of the DFT, and $g(t)$ is set to a constant (the bias), then ND is identical to the iDFT.
However, by choosing a $g(t)$ better suited to generalization and by learning the amplitudes and tuning the frequencies using backpropagation, our method is more effective at generalization than the iDFT.
$g(t)$ may be as simple as a linear equation or as complex as a combination of linear and nonlinear equations.
A discussion on the selection of $g(t)$ can be found in Section~\ref{sec:details}.

\begin{figure}[!t]
	\centering
	\includegraphics[width=3.0in]{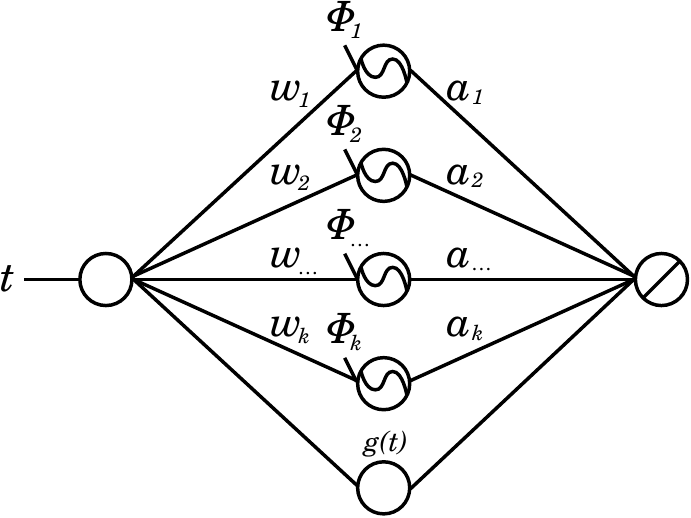}
	\caption{A diagram of the neural network model used by Neural Decomposition. For each of the $k$ sinusoid units, $w_i$ are frequencies, $\phi_i$ are phase shifts, and $a_i$ are amplitudes, where $i\in\{1\ldots k\}$. The augmentation function $g(t)$ is shown as a single unit, but it may be composed of one or more units with one or more activation functions.}
	\label{fig:gift}
\end{figure}

We use a feedforward artificial neural network with a single hidden layer to compute our function (see Figure~\ref{fig:gift}).
The hidden layer is composed of $N$ units with a sinusoid activation function and an arbitrary number of units with other activation functions to calculate $g(t)$.
The output layer is a single linear unit, so that the neural network outputs a linear combination of the units in the hidden layer.

We initialize the frequencies and phase shifts in the same way as the inverse DFT as described above.
Rather than use the actual values provided by the DFT as sinusoid amplitudes, however, we initialize them to small random values (see Section~\ref{subsec:toy_analysis} for a discussion on why).
Weights in the hidden layer associated with $g(t)$ are initialized to approximate identity, and weights in the output layer associated with $g(t)$ are randomly perturbed from zero.

\begin{figure*}[!t]
	\centering
	\includegraphics[width=6.5in]{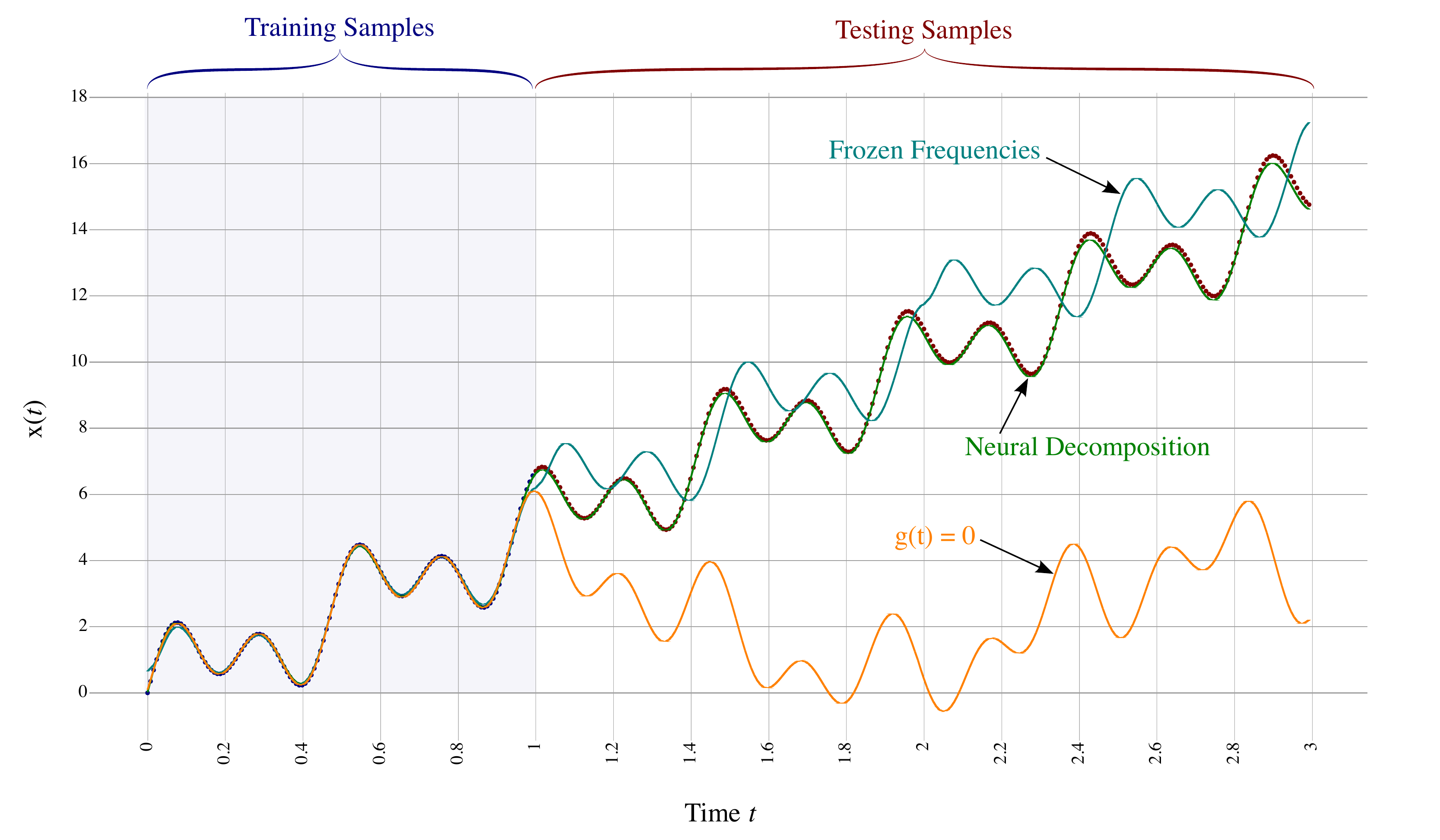}
	\caption{A comparison of Neural Decomposition with two algorithmic variations showing the importance of certain algorithm details. The data used here is the same data used in Figure~\ref{fig:idft}. The full ND model, shown in green, fits very closely to the data that was withheld during training. The cyan curve shows predictions made when the basis functions, including sinusoidal frequencies, were frozen during training. Note that the predictions are out-of-phase, indicating that training these components is essential for effective generalization. The orange curve shows predictions made without including any nonperiodic components among the basis functions, that is, setting the augmentation function $g(t) = 0$. Although the predictions exhibit the correct phase, they fail to fit with the nonperiodic trend. This shows the importance of using heterogeneous basis functions.}
	\label{fig:toy}
\end{figure*}

We train our model using stochastic gradient descent with backpropagation. This training process allows our model to learn better frequencies and phase shifts so that the sinusoid units more accurately represent the periodic components of the time-series. Because frequencies and phase shifts are allowed to change, our model can learn the true period of the underlying function rather than assuming the period is $N$. Training also tunes the weights of the augmentation function.

ND uses regularization throughout the training process to distribute weights in a manner consistent with our goal of generalization. In particular, we use $L^1$ regularization on the output layer of the network to promote sparsity by driving nonessential weights to zero. Thus, ND produces a simpler model by using the fewest number of units that still fit the training data well.

By pre-initializing the frequencies and phase shifts to mimic the inverse DFT and setting all other parameters to small values, we reduce time-series prediction to a simple regression problem. Artificial neural networks are particularly well-suited to this kind of problem, and using stochastic gradient descent with backpropagation to train it should yield a precise and accurate model.

The neural network model and training approach we use is similar to those used by Gashler and Ashmore in a previous work on time-series analysis \cite{gashler:fourier}.
Our work builds on theirs and contributes a number of improvements, both theoretically and practically.
First, we do not initialize the weights of the network using the Fourier transform.
This proved to be problematic in their work as it used periodic components to model linear and other nonperiodic parts of the training data.
By starting with weights near zero and learning weights for both periodic and nonperiodic units simultaneously, our model does not have to unlearn extraneous weights.
Second, their model required heavy regularization that favored using linear units rather than the initialized sinusoid units.
Our training process makes no assumptions about which units are more important and instead allows gradient descent to determine which components are necessary to model the data.
Third, their training process required a small learning rate (on the order of $10^{-7}$) and their network was one layer deeper than ours.
As a result, their frequencies were never tuned, their results were generally out of phase with the testing data, and their training times were very long.
Because our method facilitates the training of each frequency and allows a larger learning rate ($10^{-3}$ in our experiments), our method yields a function that is more precisely in phase with the testing data in a much shorter amount of time.
Thus, our method has simplified the complexity of the model's training algorithm, minimized its training time, and improved its overall effectiveness at time-series prediction.
The superiority of our method is demonstrated in Section~\ref{sec:results} and visualized in Figure~\ref{fig:labor}.

\subsection{Toy Problem for Justification}

\begin{figure*}[!t]
	\centering
	\includegraphics[width=3.5in]{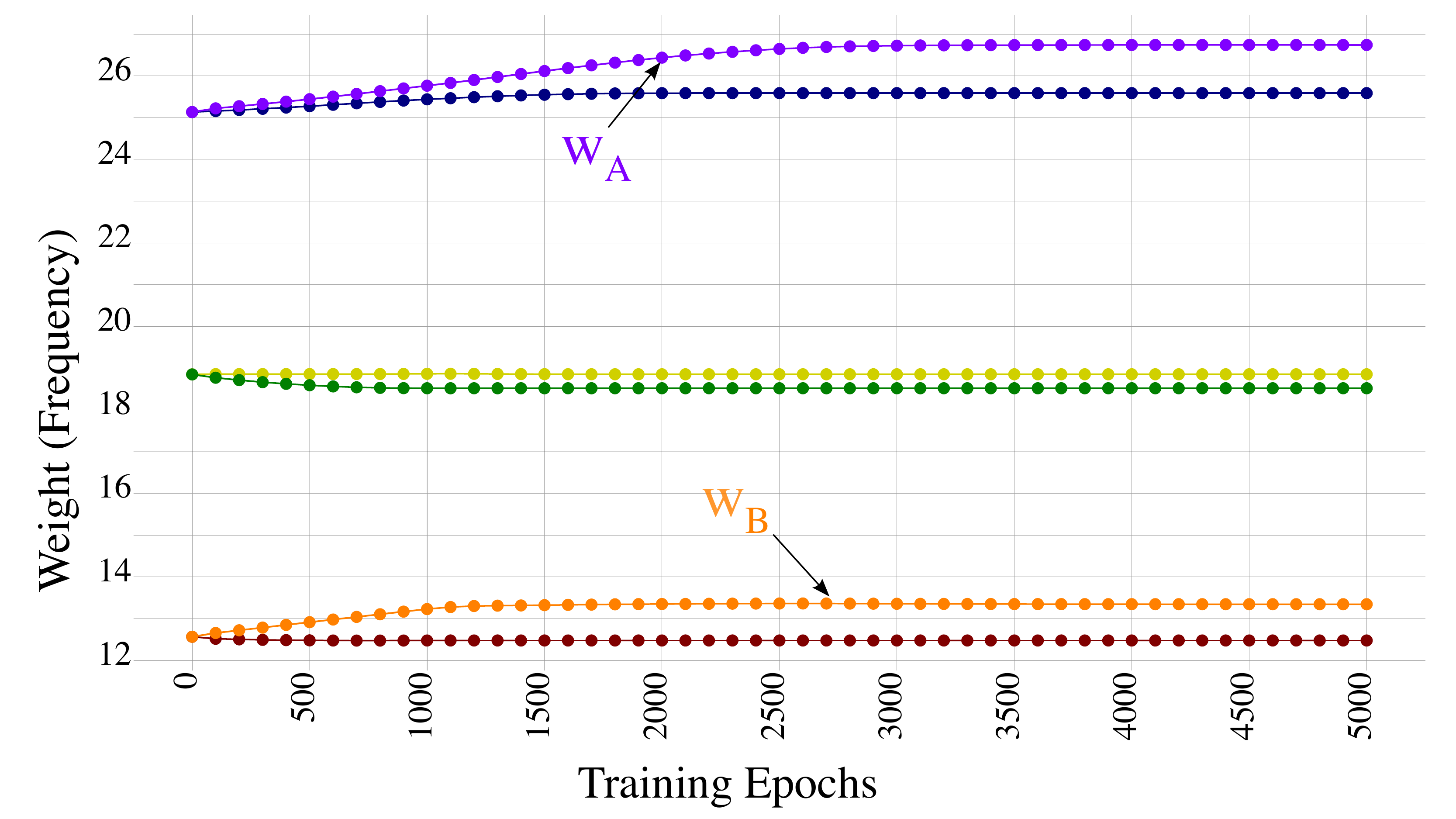}
	\includegraphics[width=3.5in]{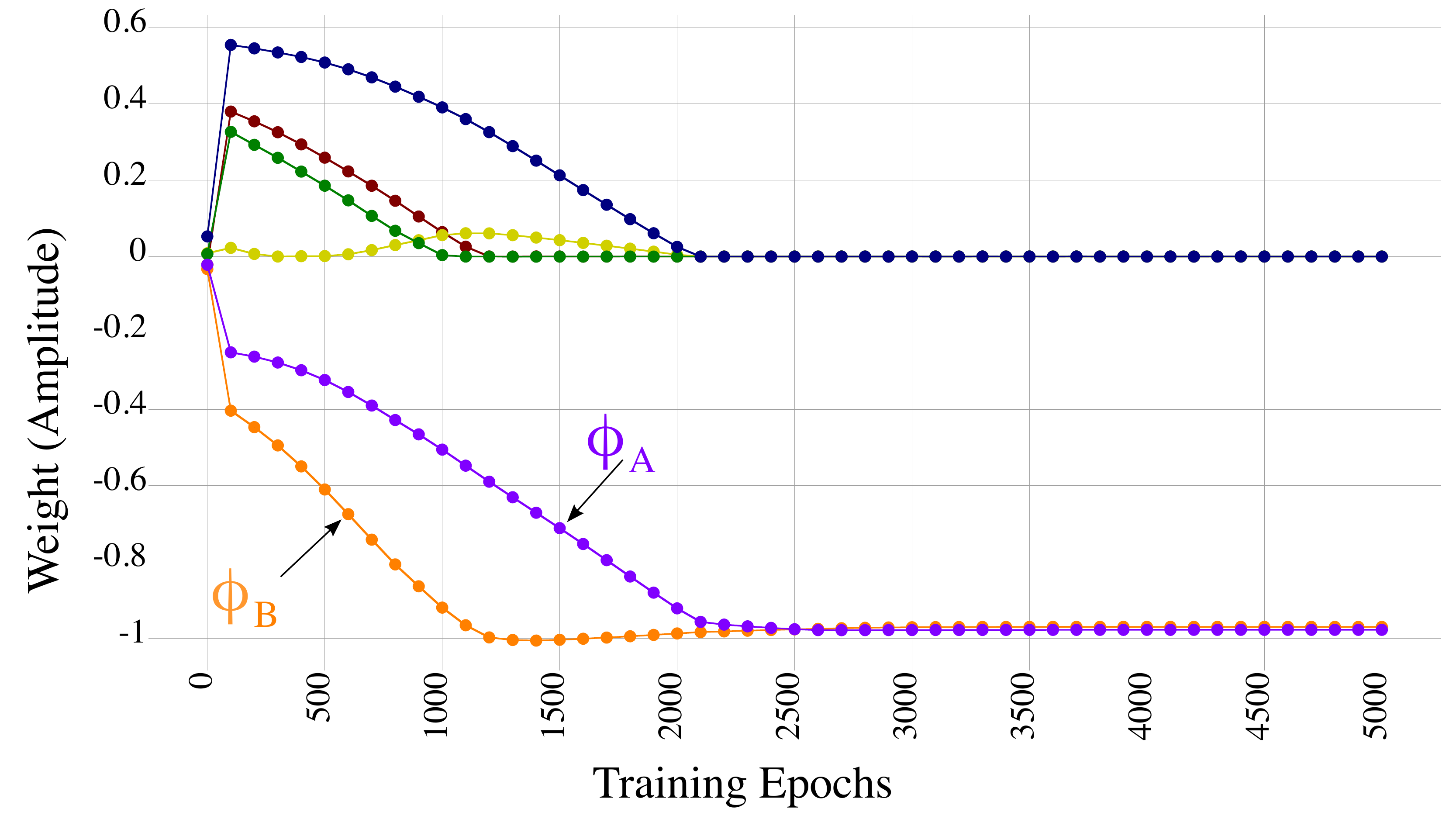}
	\caption{(Left) Frequencies of the basis functions of Neural Decomposition over time. (Right) Basis weights (amplitudes) over time on the same problem. Note that ND first tunes the frequencies (Left), then finishes adjusting the corresponding amplitudes for those sinusoids (Right) ($w_A$ corresponds to $\phi_A$ and $w_B$ corresponds to $\phi_B$). In most cases, the amplitudes are driven to zero to form a sparse representation. After the amplitudes reach zero, the frequencies are no longer modified.}
	\label{fig:weights}
\end{figure*}

\begin{figure*}[!t]
	\centering
	\includegraphics[width=3.5in]{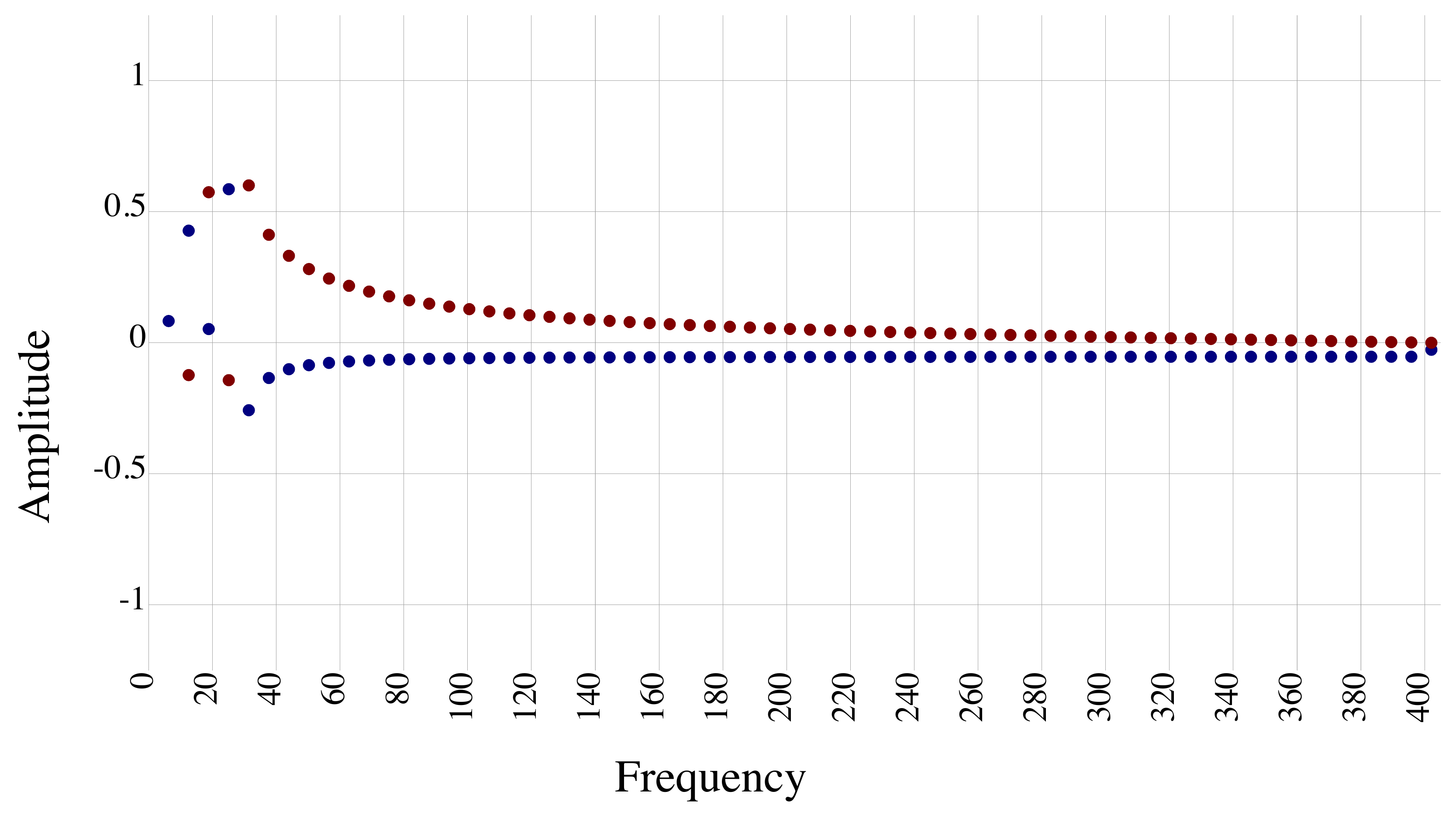}
	\includegraphics[width=3.5in]{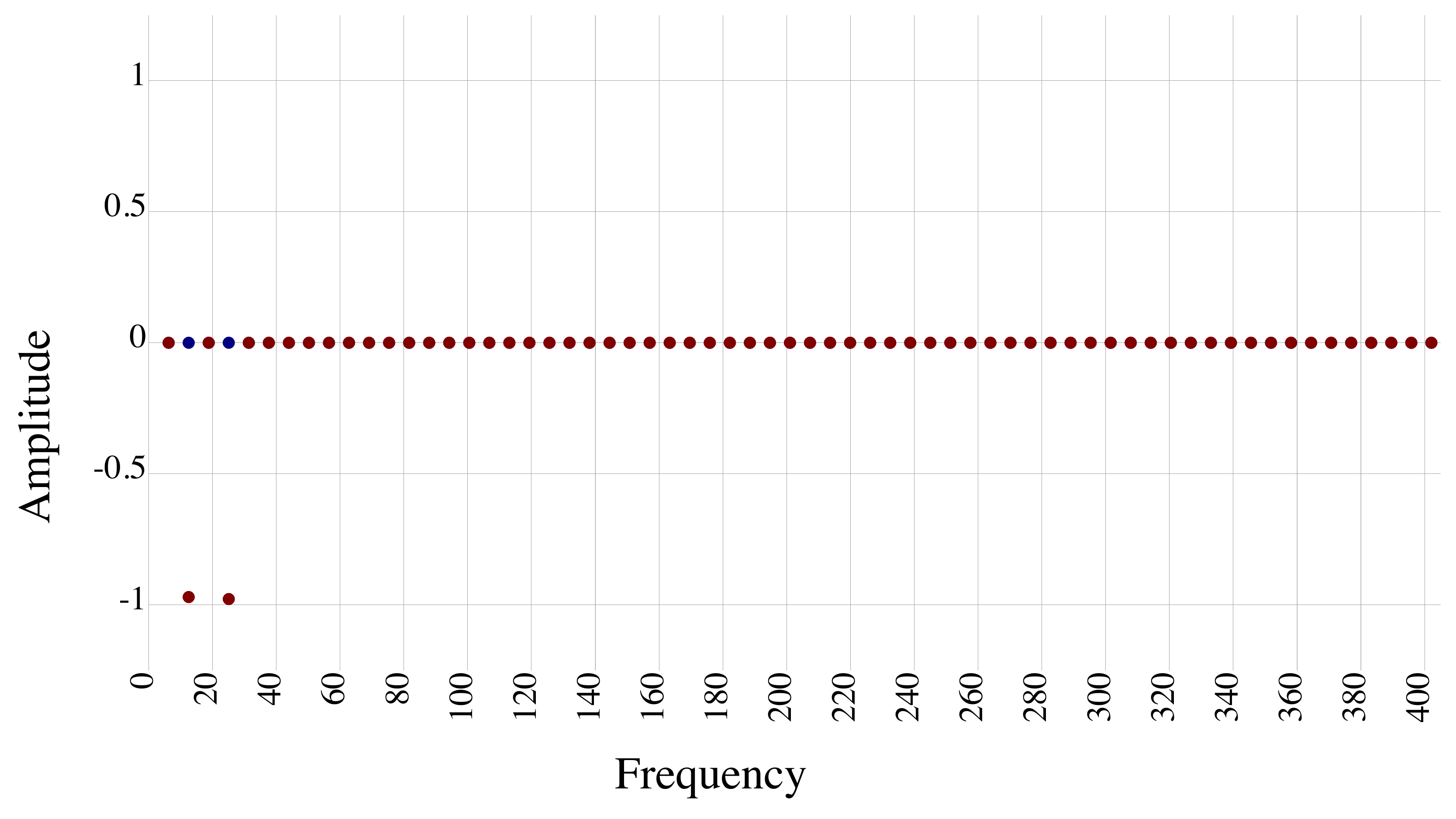}
	\caption{Frequency domain representations of the toy problem (amplitude vs frequency). (Left) Frequencies used by the iDFT. (Right) Frequencies used by ND.}
	\label{fig:freq}
\end{figure*}

Figure~\ref{fig:toy} demonstrates that flexible frequencies and an appropriate choice for $g(t)$ are essential for effective generalization. We compare three ND models using the equation $x(t) = sin(4.25 \pi t) + sin(8.5 \pi t) + 5t$ to generate time-series data. This is a sufficiently interesting toy problem because it is composed of periodic and nonperiodic functions and its period is not exactly $N$ (otherwise, the frequencies would have been multiples of $2 \pi$). We generate 128 values for $0 \leq t < 1.0$ as input and 256 values for $1.0 \leq t < 3.0$ as a validation set. Powers of two are not required, but we used powers of two in order to compare our approach with using the inverse DFT (approximated by the inverse FFT).

One of the compared ND models freezes the frequencies so that the model is unable to adjust them. Although it is able to find the linear trend in the signal, it is unable to learn the true period of the data and, as a result, makes predictions that are out of phase with the actual signal. This demonstrates that the ability to adjust the constituent parts of the output signal is necessary for effective generalization.

Another of the compared ND models has flexible frequencies, but uses no augmentation function (that is, $g(t) = 0$). This model can  learn the periodic components of the signal, but not its nonperiodic trend. It tunes the frequencies of the sinusoid units to more accurately reflect the input samples, so that it is more in phase than the second model. However, because it cannot explain the nonperiodic trend of the signal, it also uses more sinusoid units than the true underlying function requires, resulting in predictions that are not perfectly in phase. This model shows the necessity of an appropriate augmentation function for handling nonperiodicity.

The final ND model compared in Figure~\ref{fig:toy} is ND with flexible frequencies and augmentation function $g(t) = wt + b$.
As expected, it learns both the true period and the nonperiodic trend of the signal.
We therefore conclude that an appropriate augmentation function and the ability to tune components are essential in order for ND to generalize well.

\subsection{Toy Problem Analysis}
\label{subsec:toy_analysis}

In Figure~\ref{fig:weights}, we plot the weights over time of our $g(t) = wt + b$ model being trained on the toy problem.
Weights in Figure~\ref{fig:weights}(a) are the frequencies of a few of the sinusoids in the model, initialized based on the iDFT, but tuned over time to learn more appropriate frequencies for the input samples, and  weights in Figure~\ref{fig:weights}(b) are their corresponding amplitudes.
The training process tunes frequencies $w_A$ and $w_B$ to more accurately reflect the period of the underlying function and adjusts the corresponding amplitudes $\phi_A$ and $\phi_B$ so that only the sinusoids associated with these amplitudes are used in the trained model and all other amplitudes are driven to zero.
This demonstrates that ND tunes frequencies it needs and learns amplitudes as we hypothesized.
It is also worth noting that after the first 2500 training epochs, no further adjustments are made to the weights.
This suggests that ND is robust against overfitting, at least in some cases, as the ``extra'' training epochs did not result in a worse prediction.

\begin{figure*}[!t]
	\centering
	\includegraphics[width=6.5in]{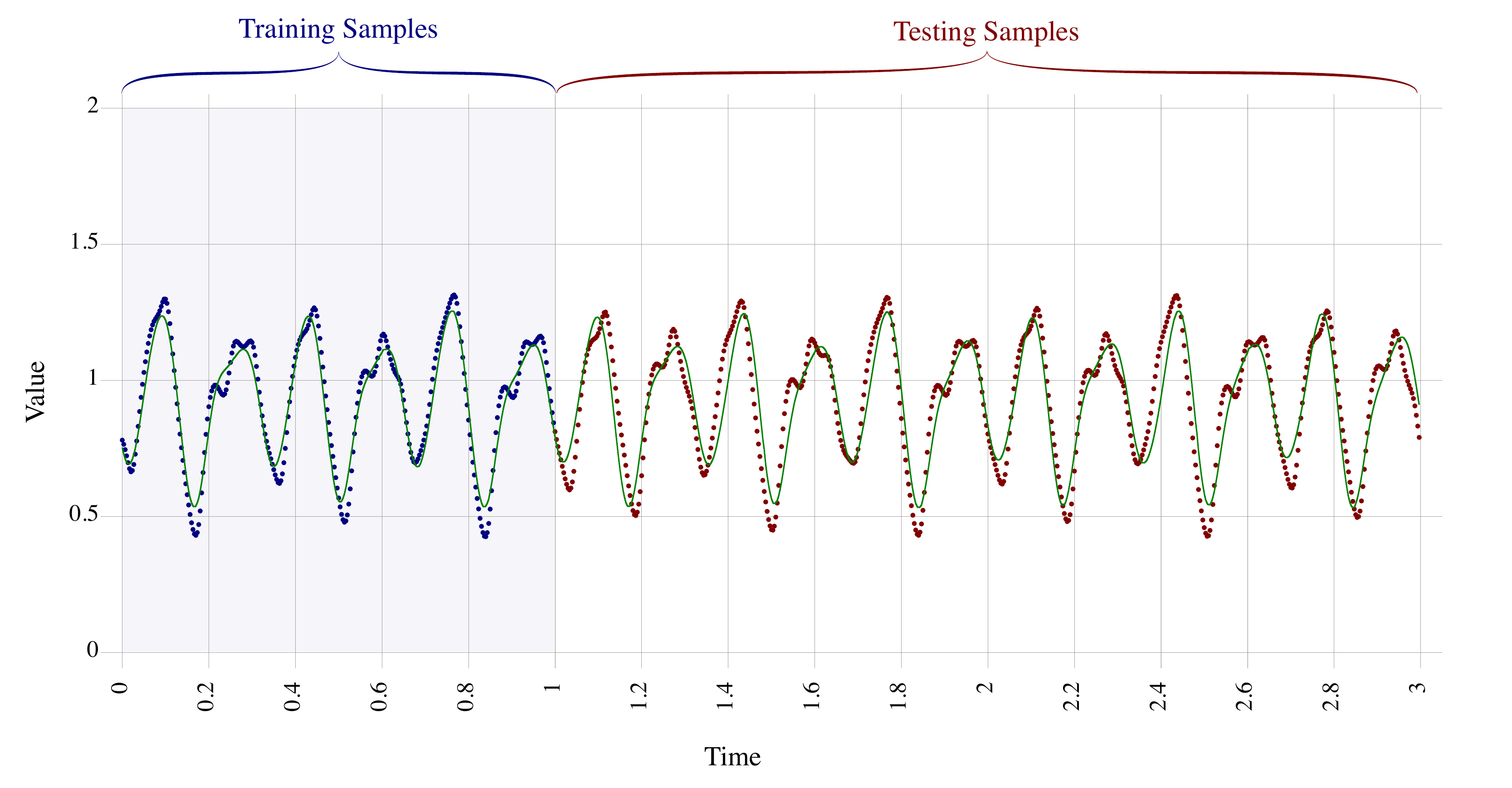}
	\caption{Neural Decomposition on the Mackey-Glass series. Although it does not capture all the high-frequency fluctuations in the data, our model predicts the location and height of each peak and valley in the series with a high degree of accuracy.}
	\label{fig:mackey}
\end{figure*}

Gashler and Ashmore utilized the FFT to initialize the sinusoid amplitudes so that the neural network immediately resembled the iDFT \cite{gashler:fourier}. Using the DFT in this way yields an unnecessarily complex model in which nearly every sinusoid unit has a nonzero amplitude, either because it uses periodic functions to model the nonperiodic signal or because it has fixed frequencies and so uses a range of frequencies to model the actual frequencies in the signal \cite{silvescu:fourier}. Consequently, the training process required heavy regularization of the sinusoid amplitudes in order to shift the weight to the simpler units (see Section~\ref{subsec:fourier_nn}). Training from this initial point often fell into local optima, as such a model was not always able to unlearn superfluous sinusoid amplitudes.

Figure~\ref{fig:freq} demonstrates why using amplitudes provided by the Fourier transform is a poor initialization point.
The actual underlying function only requires two sinusoid units (found by ND), but the Fourier transform uses every sinusoid unit available to model the linear trend in the toy problem.
Instead of tuning two amplitudes, a model initialized with the Fourier transform has to tune every amplitude and is therefore far more likely to fall into local optima.

\begin{figure*}[!t]
	\centering
	\includegraphics[width=6.5in]{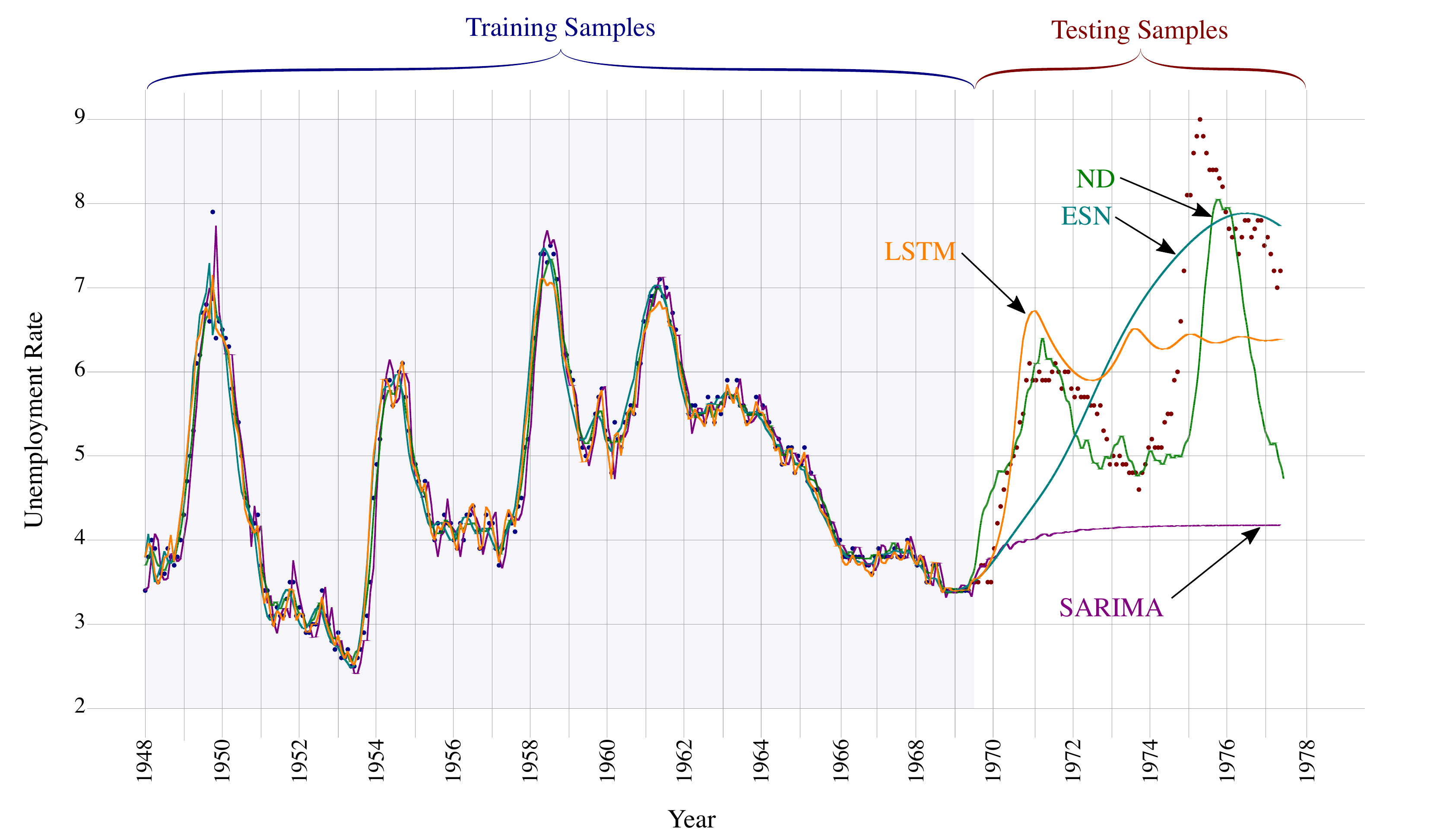}
	\caption{A comparison of the four best predictive models on the monthly unemployment rate in the US.
	Blue points represent training samples from January 1948 to June 1969 and red points represent testing samples from July 1969 to December 1977.
	SARIMA, shown in magenta, correctly predicted a rise in unemployment but underestimated its magnitude, and did not predict the shape of the data well.
	ESN, shown in cyan, predicted a reasonable mean, but did not capture the dynamics of the data.
	LSTM, shown in orange, predicted the first peak in the data, but leveled off to predict only the mean.
	Only ND, shown in green, successfully predicted both the depth and approximate shape of the surge in unemployment, followed by another surge in unemployment that followed.}
	\label{fig:labor}
\end{figure*}

ND, by contrast, does not use the FFT.
Sinusoid amplitudes (the weights feeding into the output layer) and all output-layer weights associated with $g(t)$ are initialized to small random values.
This allows the neural network to learn the periodic and nonperiodic components of the signal simultaneously.
Not only does this avoid unnecessary ``unlearning'' of the extra weights used by the DFT, but also avoids getting stuck in the local optima represented by the DFT weights.
Without the hindrance of having to unlearn part of the DFT, the training process is able to find more optimal values for these weights.
Figure~\ref{fig:freq} shows a comparison of our trained model with the frequencies used by the iDFT, omitting the linear component learned by ND.

\subsection{Chaotic Series}
\label{subsec:mackey}

In addition to the toy problem, we applied ND to the Mackey-Glass series as a proof-of-concept.
This series is known to be chaotic rather than periodic, so it is an interesting test for our approach that decomposes the signal as a combination of sinusoids.
Results with this data are shown in Figure~\ref{fig:mackey}.
The blue points on the left represent the training sequence, and the red points on the right half represent the testing sequence.
All testing samples were withheld from the model, and are only shown here to illustrate the effectiveness of the model in anticipating future samples.
The green curve represents the predictions of the trained model.
The series predicted by Neural Decomposition exhibits shapes similar to those in the test data, and has an RMSE of 0.086.
Interestingly, neither the shapes in the test data nor those exhibited within the model are strictly repeating.
This occurs because the frequencies of the sinusoidal basis functions that ND uses to represent its model may be tuned to have frequencies with no small common multiple, thus creating a signal that does not repeat for a very long time.
Our model does not capture all the high-frequency fluctuations, but it is able to approximate the general shape and some of the dynamics of the chaotic series.

To determine whether Neural Decomposition merely predicts a periodic function, we tried our experiment again but set $g(t) = 0$ rather than using nonlinear, nonperiodic components for $g(t)$.
We found that with these changes, our model was unable to capture the subtle dynamics of the Mackey-Glass series.
As in the toy problem, omitting $g(t)$ resulted in poorer predictions, and the resulting predictions had an RMSE of 0.14 (a 63\% increase in error).
This indicates that ND does more than predict a strictly periodic function, and is able to capture at least some of the nonlinear dynamics in some chaotic systems.

Although preliminary tests on the toy problem and the Mackey-Glass series were favorable to Neural Decomposition, not all of our tests were as successful.
In particular, we applied ND to another chaotic series: samples from the Lorenz-63 model.
We found that ND was unable to effectively model the dynamics of this chaotic system.
This seems to indicate that although ND does well with some problems, it should not be expected to anticipate all the subtle variations that occur in chaotic systems.

\begin{figure*}[!t]
	\centering
	\includegraphics[width=6.5in]{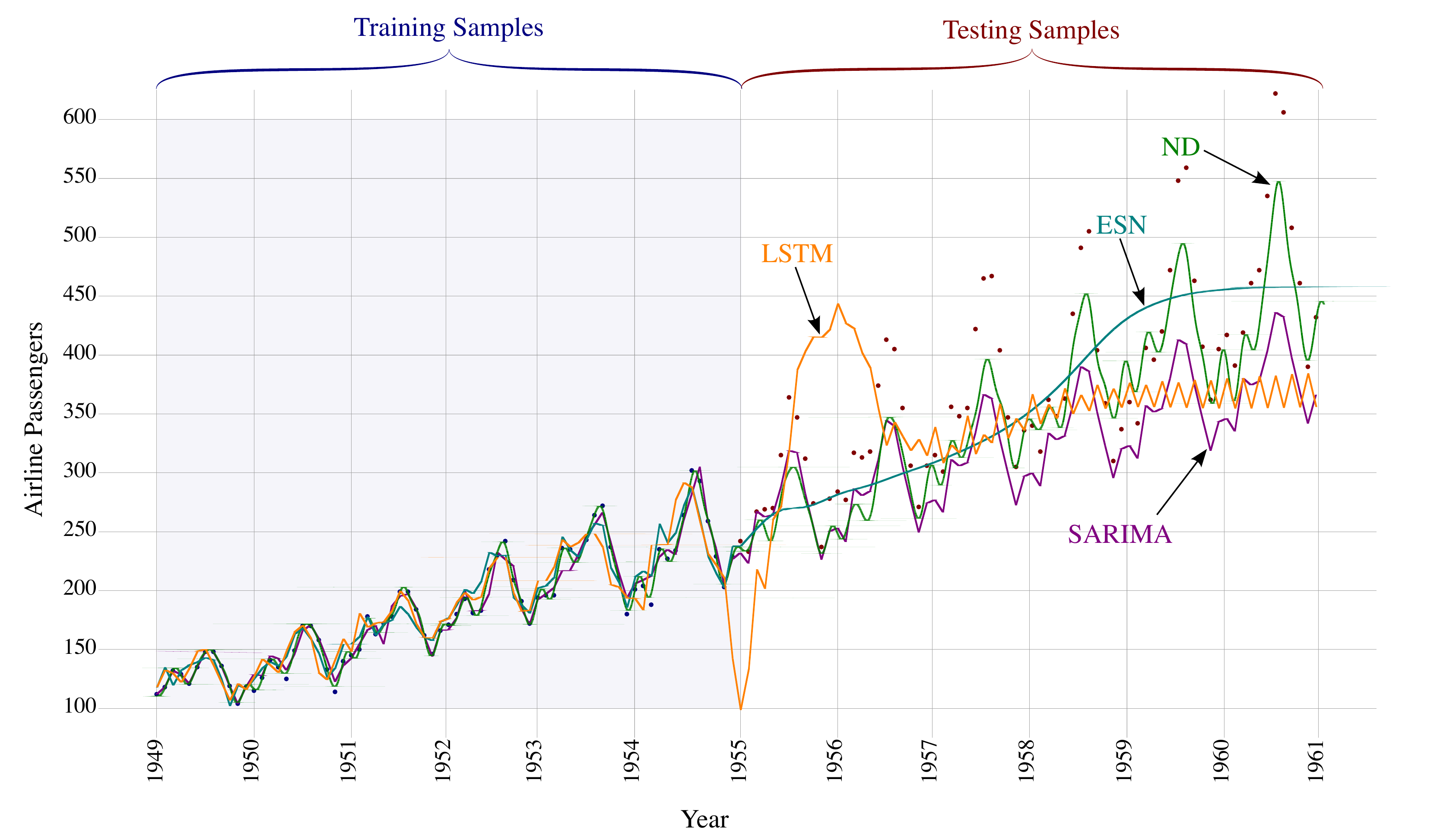}
	\caption{A comparison of the four best predictive models on monthly totals of international airline passengers from January 1949 to December 1960 \cite{chatfield:analysis}.
	Blue points represent the 72 training samples from January 1949 to December 1954 and red points represent the 72 testing samples from January 1955 to December 1960.
	SARIMA, shown in magenta, learns the trend and general shape of the data.
	ESN, shown in cyan, predicts a mean but does not capture the dynamics of the actual data.
	LSTM, shown in orange, predicts a valley and a peak that did not actually occur, followed by a poor estimation of the mean that suggests that it was unable to learn the seasonality of the data.
	ND, shown in green, learns the trend, shape, and growth better than the other compared models.}
	\label{fig:airline}
\end{figure*}

\begin{figure*}[!t]
	\centering
	\includegraphics[width=6.5in]{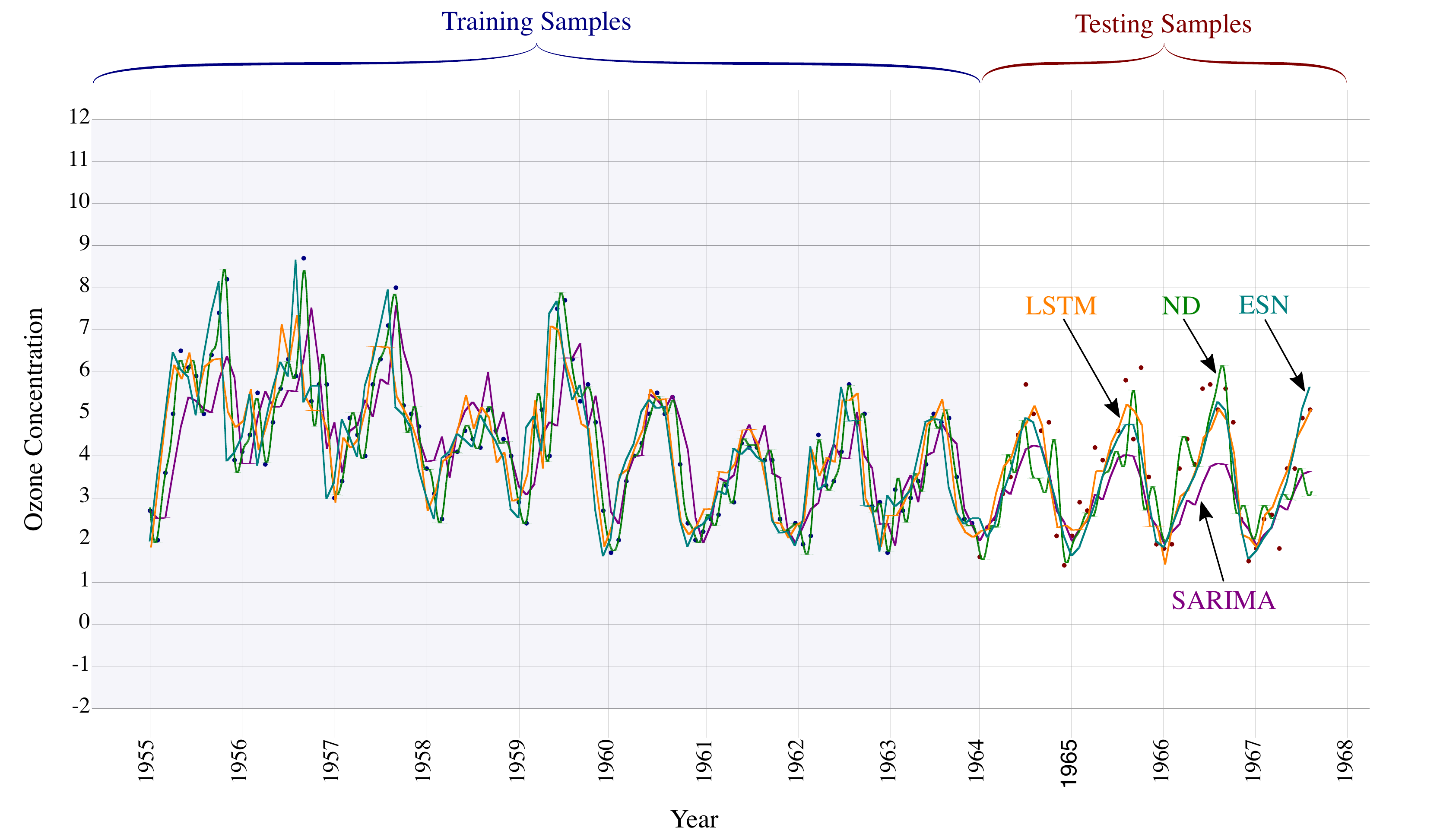}
	\caption{A comparison of the four best predictive models on monthly ozone concentration in downtown Los Angeles from January 1955 to August 1967 \cite{hipel:modeling}.
	Blue points represent the 152 training samples from January 1955 to December 1963 and red points represent the 44 testing samples from January 1964 to August 1967.
	The compared models include SARIMA, ESN, LSTM, and ND.
	All four of these models perform well on this problem.
	Both LSTM (shown in orange) and ESN (shown in cyan) predict with slightly higher accuracy compared to ND.
	ND, shown in green, has slightly higher accuracy compared to SARIMA (shown in magenta).
	ARIMA, SVR, and Gashler and Ashmore's model all performed poorly on this problem; rather than include them in this graph, their errors have been reported in Table~\ref{table:mape} and Table~\ref{table:rmse}.}
	\label{fig:ozone}
\end{figure*}

\section{Implementation Details}
\label{sec:details}

In this section, we provide a more detailed explanation of our approach. A high level description of Neural Decomposition can be found in Section~\ref{sec:high-level}. For convenience, an implementation of Neural Decomposition is included in the Waffles machine learning toolkit \cite{gashler2011jmlr}.

\subsection{Topology}

We use a feedforward artificial neural network as the basis of our model. For an input of size $N$, the neural network is initialized with two layers: $1 \rightarrow m$ and $m \rightarrow 1$, where $m = N + |g(t)|$ and $|g(t)|$ denotes the number of nodes required by $g(t).$ The first $N$ nodes in the hidden layer have the sinusoid activation function, $sin(t)$, and the rest of the nodes in the hidden layer have other activation functions to compute $g(t)$.

The augmentation function $g(t)$ can be made up of any number of nodes with one or more activation functions.
For example, it could be made up of linear units for learning trends and sigmoidal units to fit nonperiodic, nonlinear irregularities.
Gashler and Ashmore have suggested that softplus units may yield better generalizing predictions compared to standard sigmoidal units \cite{gashler:fourier}.
In our experiments, we used a combination of linear, softplus, and sigmoidal nodes for $g(t)$.
The network tended to only use a single linear node, which may suggest that the primary benefit of the augmentation function is that it can model linear trends in the data.
Softplus and sigmoidal units tended to be used very little or not at all by the network in the problems we tested, but intuitively it seems that nonlinear activation functions could be useful in some cases.

\subsection{Weight Initialization}

The weights of the neural network are initialized as follows.
Let each of the $N$ sinusoid nodes in the hidden layer, indexed as $k$ for $0 \leq k < N$, have a weight $w_k$ and bias $\phi_k$.
Let each $w_k$ represent a frequency and be initialized to $2 \pi \lfloor k/2 \rfloor$.
Let each $\phi_k$ represent a phase shift.
For each even value of $k$, let $\phi_k$ be set to $\pi/2$ to transform $sin(t + \phi_k)$ to $cos(t)$.
For each odd value of $k$, let $\phi_k$ be set to $\pi$ to transform $sin(t + \phi_k)$ to $-sin(t)$.
A careful comparison of these initialized weights with Equation~\ref{eq:idft} shows that these are identical to the frequencies and phase shifts used by the iDFT, except for a missing $1/N$ term in each frequency, which is absorbed in the input preprocessing step (see Subsection~\ref{subsec:preprocessing}).

All weights feeding into the output unit are set to small random values. At the beginning of training, therefore, the model will predict something like a flat line centered at zero. As training progresses, the neural network will learn how to combine the hidden layer units to fit the training data.

Weights in the hidden layer associated with the augmentation function are initialized to approximate the identity function. For example, in $g(t) = wt + b$, $w$ is randomly perturbed from 1 and $b$ is randomly perturbed near 0. Because the output layer will learn how to use each unit in the hidden layer, it is important that each unit be initialized in this way.

\subsection{Input Preprocessing}
\label{subsec:preprocessing}

Before training begins, we preprocess the input data to facilitate learning and prevent the model from falling into a local optimum.
First, we normalize the time associated with each sample so that the training data lies between 0 (inclusive) and 1 (exclusive) on the time axis.
If there is no explicit time, equally spaced values between 0 and 1 are assigned to each sample in order.
Predicted data points will have a time value greater than or equal to 1 by this new scale.
Second, we normalize the values of each input sample so that all training data is between 0 and 10 on the y axis.

This preprocessing step serves two purposes. First, it absorbs the $1/N$ term in the frequencies by transforming $t$ into $t/N$, which is why we were able to omit the $1/N$ term from our frequencies in the weight initialization step. Second, and more importantly, it ensures that the data is appropriately scaled so that the neural network can learn efficiently. If the data is scaled too large on either axis, training will be slow and susceptible to local optima. If the data is scaled too small, on the other hand, the learning rate of the machine will cause training to diverge and only use linear units and low frequency sinusoids.

In some cases, it is appropriate to pass the input data through a filter. For example, financial time-series data is commonly passed through a logarithmic filter before being presented for training, and outputs from the model can then be exponentiated to obtain predictions. We use this input preprocessing method in two of our experiments where we observe an underlying exponential growth in the training data.

\subsection{Regularization}

Regularization is essential to the training process. Prior to each sample presentation, we apply regularization on the output layer of the neural network. Even though we do not initialize sinusoid amplitudes using the DFT, the network is quickly able to learn how to use the initialized frequencies to perfectly fit the input samples. Without regularizing the output layer, training halts as soon as the model fits the input samples, because the measurable error is near zero. By relaxing the learned weights, regularization allows our model to redistribute its weight over time. We find that regularization amount is especially important; too much prevented our model from learning, but too little caused our model to fall into local optima. In our experiments, setting the regularization term to $10^{-2}$ avoided both of these potential pitfalls.

Another important function of regularization in ND is to promote sparsity in the network, so that the redistribution of weight produces as simple a model as the input samples allow. We use $L^1$ regularization for this reason. Usually, the trained model does not require all $N$ sinusoid nodes in order to generalize well, and this type of regularization enables the network to automatically discard unnecessary nodes by driving their amplitudes to zero. $L^2$ regularization is not an acceptable substitute in this case, as it would distribute the weights evenly throughout the network and could, like the DFT, try to use several sinusoid nodes to model what would more appropriately be modeled by a single node with a nonperiodic activation function.

It is worth noting that we only apply regularization to the output layer of the neural network. Any regularization that might occur in the hidden layer would adjust sinusoid frequencies before the output layer could learn sinusoid amplitudes. By allowing weights in the hidden layer to change without regularization, the network has the capacity to adjust frequencies but is not required to do so.

Backpropagation with stochastic gradient descent tunes the weights of the network and accomplishes the redistribution of weights that regularization makes possible. In our experiments, we use a learning rate of $10^{-3}$.

\begin{figure*}[!t]
	\centering
	\includegraphics[width=6.5in]{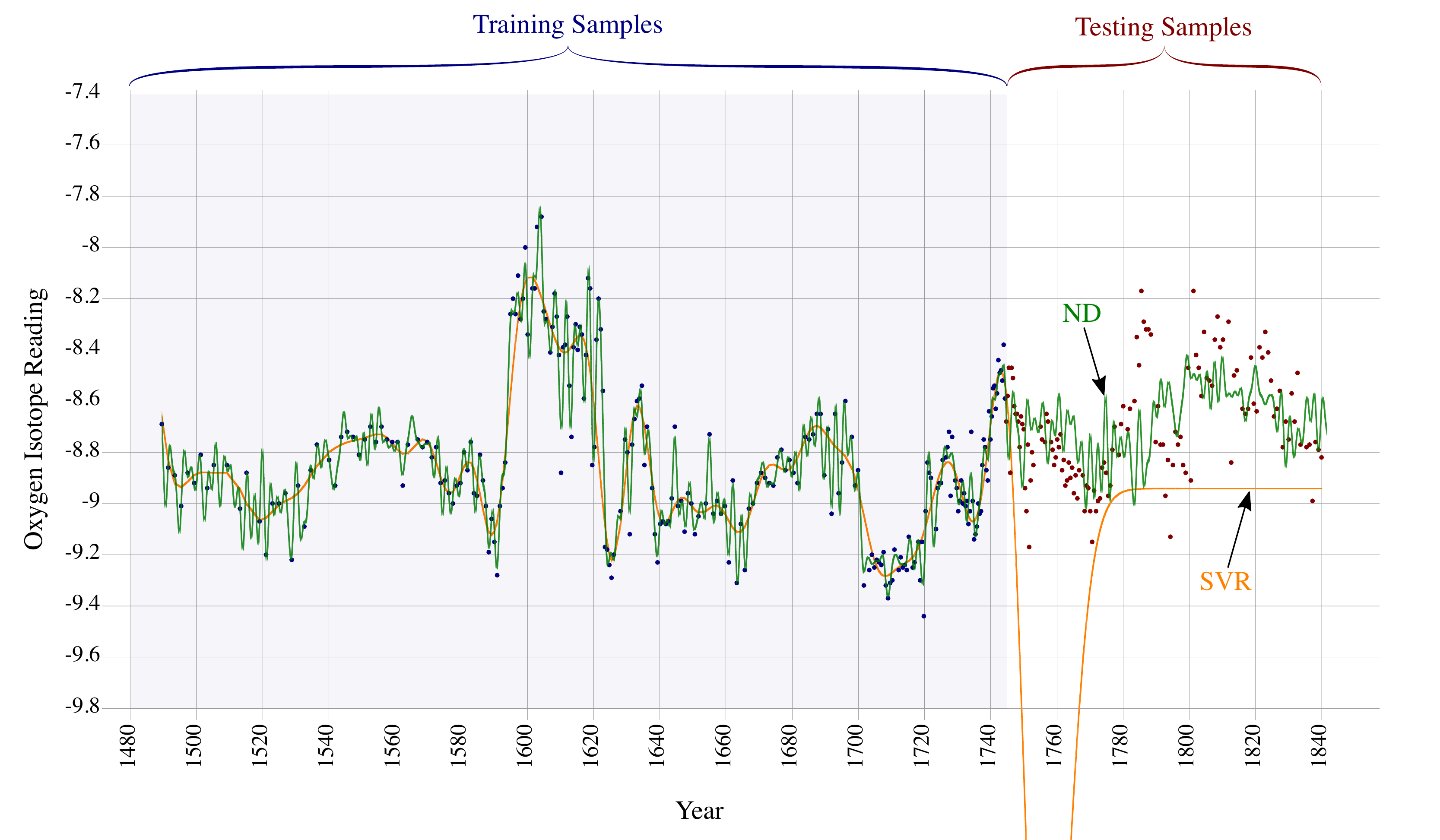}
	\caption{A comparison of two predictive models on a series of oxygen isotope readings in speleothems in India from 1489 AD to 1839 AD \cite{Sinha:2015aa}.
	Blue points represent the 250 training samples from July 1489 to April 1744 and red points represent the 132 testing samples from August 1744 to December 1839.
	Because this time-series is irregularly sampled (the time step between samples is not constant), only SVR and ND could be applied to it.
	SVR, shown in orange, does not perform well, but predicts a steep drop in value that does not actually occur in the testing data, followed by a flat line.
	ND, shown in green, performs well, capturing the general shape of the testing samples.}
	\label{fig:o18}
\end{figure*}

\section{Validation}
\label{sec:results}

In this section, we report results that validate the effectiveness of Neural Decomposition.
In each of these experiments, we used an ND model with an augmentation function made up of ten linear units, ten softplus units, and ten sigmoidal units.
It is worth noting that $g(t)$ is under no constraint to consist only of these units; it could include other activation functions or only contain a single linear node to capture trend information.
We use a regularization term of $10^{-2}$ and a learning rate of $10^{-3}$ in every experiment to demonstrate the robustness of our approach; we did not tune these meta-parameters for each experiment.

In our experiments, we compare ND with LSTM, ESN, ARIMA, SARIMA, SVR, Gashler and Ashmore's model \cite{gashler:fourier}.
We used PyBrain's implementation of LSTM networks \cite{pybrain2010jmlr} with one input neuron, one output neuron, and one hidden layer.
We implemented a grid-search to find the best hidden layer size for the LSTM network for each problem and used PyBrain's RPROP- algorithm to train the network.
We used Luko{\v s}evi{\v c}ius' implementation of ESN \cite{lukosevicius2012esn} and implemented a grid-search to find the best parameters for each problem.
We used the R language implementation for ARIMA, SARIMA, and SVR \cite{r:language}.
For the ARIMA models, we used a variation of the \texttt{auto.arima} method that performs a grid-search to find the best parameters for each problem.
For SVR, we used the \texttt{tune.svm} method, which also performs a grid-search for each problem.
Although these methods select the best models based on the amount of error calculated using the training samples, the grid-search is a very slow process.
Gashler and Ashmore's model did not require a grid-search for parameters because it has a default set of parameters that are automatically tuned during the training process.
With ND, no problem-specific parameter tuning was performed.

In each figure, the blue points in the shaded region represent training samples and the red points represent withheld testing samples.
The curves on the graph represent the predictions made by the four models that made the most accurate predictions (only two models are shown in the fourth experiment because only two models could be applied to an irregularly sampled time-series).
The actual error for each model's prediction is reported for all experiments and all models in Table~\ref{table:mape} and Table~\ref{table:rmse}.

The LSTM network tended to fall into local optima, and was thus extremely sensitive to the random seed.
Running the same experiment with LSTM using a different random seed yielded very different results.
In each experiment, therefore, we tried the LSTM model 100 times for each different topology tested in our grid-search and selected the result with the highest accuracy to present for comparison with ND.
Conversely, ND consistently made approximately identical predictions when run multiple times, regardless of the random seed.


In our first experiment, we demonstrated the effectiveness of ND on real-world data compared to widely used techniques in time-series analysis and forecasting.
We trained our model on the unemployment rate from 1948 to 1969 as reported by the U.S. Bureau of Labor Statistics, and predicted the unemployment rate from 1969 to 1977.
These results are shown in Figure~\ref{fig:labor}.
Blue points on the left represent the 258 training samples from January 1948 to June 1969, and red points on the right represent the 96 testing samples from July 1969 to December 1977.
The four curves represent predictions made by ND (green), LSTM (orange), ESN (cyan), and SARIMA (magenta); ARIMA, SVR, and Gashler and Ashmore's model yielded poorer predictions and are therefore omitted from the figure.
Grid-search found ARIMA(3,1,2) and ARIMA(1,1,2)(1,0,1)[12] for the ARIMA and SARIMA models, respectively.
ARIMA, not shown, did not predict the significant rise in unemployment.
SARIMA, shown in magenta, did correctly predict a rise in unemployment, but underestimated its magnitude, and did not predict the shape of the data well.
SVR, not shown, correctly predicted that unemployment would rise, then fall again. However, it also underestimated the magnitude.
ESN, shown in cyan, predicted a reasonable mean value for the general increase in unemployment, but failed to capture the dynamics of the actual data.
The best LSTM network topology, found by grid-search, had a hidden layer with 16 neurons. LSTM, shown in orange, predicted the first peak in the data, but leveled off to predict only the mean.
Gashler and Ashmore's model, not shown, predicted the rise and fall in unemployment, but underestimated its magnitude and the model's predictions significantly diverge from the subsequent testing samples. It is also worth noting that Gashler and Ashmore's model took about 200 seconds to train compared to ND, which took about 30 seconds to train.

Results with Neural Decomposition (ND) are shown in green. ND successfully predicted both the depth and approximate shape of the surge in unemployment. Furthermore, it correctly anticipated another surge in unemployment that followed. ND did a visibly better job of predicting the nonlinear trend much farther into the future.

Our second experiment demonstrates the versatility of Neural Decomposition by applying to another real-world dataset: monthly totals of international airline passengers as reported by Chatfield \cite{chatfield:analysis}.
We use the first six years of data (72 samples) from January 1949 to December 1954 as training data, and the remaining six years of data (72 samples) from January 1955 to December 1960 as testing data.
The training data is preprocessed through a $log(x)$ filter and the outputs are exponentiated to obtain the final predictions.
As in the first experiment, we compare our model with LSTM, ESN, ARIMA, SARIMA, SVR, and the model proposed by Gashler and Ashmore.
The predictions of the four most accurate models (ND, LSTM, ESN, and SARIMA) are shown in Figure~\ref{fig:airline}; ARIMA, SVR, and Gashler and Ashmore's model yielded poorer predictions and are therefore omitted from the figure.
SVR, not shown, predicts a flat line after the first few time steps and generalizes the worst out of the four predictive models.
The ARIMA model found by grid-search was ARIMA(2,1,3). ARIMA, not shown, was able to learn the trend, but failed to capture any of the dynamics of the signal.
Grid-search found ARIMA(1,0,0)(1,1,0)[12] for the SARIMA model. Both SARIMA (shown in magenta) and ND (shown in green) are able to accurately predict the shape of the future signal, but ND performs better.
Unlike SARIMA, ND learns that the periodic component gets bigger over time.
Gashler and Ashmore's model makes meaningful predictions for a few time steps, but appears to diverge after the first predicted season.
ESN, shown in cyan, performs similarly to the ARIMA model, only predicting the trend and failing to capture seasonal variations.
The LSTM network, with a hidden layer of size 64 found by grid-search, failed to capture any meaningful seasonality in the training data. Instead, LSTM immediately predicted a valley and a peak that did not actually occur, followed by a poor estimation of the mean.

The third experiment uses the monthly ozone concentration in downtown Los Angeles as reported by Hipel \cite{hipel:modeling}.
Nine years of monthly ozone concentrations (152 samples) from January 1955 to December 1963 are used as training samples, and the remaining three years and eight months (44 samples) from January 1964 to August 1967 are used as testing samples.
The training data, as in the second experiment, is preprocessed through a $log(x)$ filter and output is exponentiated to obtain the final predictions.
Figure~\ref{fig:ozone} compares the SARIMA, ESN, LSTM, and ND models on this problem; ARIMA, SVR, and Gashler and Ashmore's model yielded poorer predictions and are therefore omitted from the figure.
The ARIMA and SARIMA models found by grid-search were ARIMA(2,1,2) and ARIMA(1,1,1)(1,0,1)[12], respectively.
ARIMA and SVR resulted in flat-line predictions with a high amount of error, and Gashler and Ashmore's model diverged in training and yielded unstable predictions.
SARIMA (shown in magenta), ESN (shown in cyan), LSTM (shown in orange), and ND (shown in green), on the other hand, all forecast future samples well.
LSTM and ESN yielded the most accurate predictions, but SARIMA and ND yielded good results as well.

Our fourth experiment demonstrates that ND can be used on irregularly sampled time-series.
We use a series of oxygen isotope readings in speleothems in a cave in India from 1489 AD to 1839 AD as reported by Sinha et. al \cite{Sinha:2015aa}.
Because the time intervals between adjacent samples is not constant (the interval is about 1.5 years on average, but fluctuates between 0.5 and 2.0 years), only ND and SVR models can be applied.
ARIMA, SARIMA, Gashler and Ashmore's model, ESN, and LSTM cannot be applied to irregular time-series because they assume a constant time interval between adjacent samples; these five models are therefore not included in this experiment.
Figure~\ref{fig:o18} shows the predictions of ND and SVR.
Blue points on the left represent the 250 training samples from July 1489 to April 1744, and red points on the right represent the 132 testing samples from August 1744 to December 1839.
SVR, shown in orange, predicts a steep drop in value that does not exist in the testing data.
ND, shown in green, accurately predicts the general shape of the testing data.

Table~\ref{table:mape} presents an empirical evaluation of each model for the four real-world experiments. We use the mean absolute percent error (MAPE) as our error metric for comparisons \cite{lippi:forecasting}. MAPE for a set of predictions is defined by the following function, where $x_t$ is the actual signal value (i.e. it is an element  of the set of testing samples) and $x(t)$ is the predicted value:

\begin{equation}
	MAPE = \frac{1}{n} \sum_{t=1}^{n}{\left| \frac{x_t - x(t)}{x_t} \right|}
\end{equation}

Using MAPE, we compare Neural Decomposition to ARIMA, SARIMA, SVR with a radial basis function, Gashler and Ashmore's model, ESN, and LSTM.
We found that on the unemployment rate problem (Figure~\ref{fig:labor}), ND yielded the best model, followed by LSTM and ESN.
On the airline problem (Figure~\ref{fig:airline}), ND performed significantly better than all of the other approaches.
On the ozone problem (Figure~\ref{fig:ozone}), LSTM and ESN were the best models, but ND and SARIMA also performed well.
On the oxygen isotope problem (Figure~\ref{fig:o18}), ND outperformed SVR, which was the only other model that could be applied to the irregular time-series.
Table~\ref{table:mape} presents the results of our experiments, and Table~\ref{table:rmse} presents the same data using the root mean square error (RMSE) metric instead of MAPE.
In each problem, the accuracy of the best algorithm is shown in bold.

\begin{table}
	\centering
	\caption{Mean absolute percent error (MAPE) on the validation problems for ARIMA, SARIMA, SVR, Gashler and Ashmore, ESN, LSTM, and ND. Best result (smallest error) for each problem is shown in \textbf{bold}.}
	\begin{tabular}{| l || l | l | l | l |}
		\hline
		\textbf{Model}	& \textbf{Labor}		& \textbf{Airline}	& \textbf{Ozone}	& \textbf{Speleothem} \\ \hline
		ARIMA			& 39.42\%				& 12.34\%			& 39.50\%			& N/A \\
		SARIMA			& 29.69\%				& 13.33\%			& 22.71\%			& N/A \\
		SVR				& 25.14\%				& 47.04\%			& 49.53\%			& 8.50\% \\
		Gashler/Ashmore	& 34.38\%				& 19.89\%			& 77.19\%			& N/A \\
		ESN				& 15.73\%				& 12.05\%			& \textbf{16.15\%}	& N/A \\
		LSTM			& 14.63\%				& 18.95\%			& 16.52\%			& N/A \\
		ND				& \textbf{10.89\%}		& \textbf{9.52\%}	& 21.59\%			& \textbf{1.89\%} \\ \hline
	\end{tabular}
	\label{table:mape}
\end{table}

\begin{table}
	\centering
	\caption{Root mean square error (RMSE) on the validation problems for ARIMA, SARIMA, SVR, Gashler and Ashmore, ESN, LSTM, and ND. Best result (smallest error) for each problem is shown in \textbf{bold}.}
	\begin{tabular}{| l || l | l | l | l |}
		\hline
		\textbf{Model}	& \textbf{Labor}	& \textbf{Airline}	& \textbf{Ozone}	& \textbf{Speleothem} \\ \hline
		ARIMA			& 2.97				& 75.32				& 1.33				& N/A \\
		SARIMA			& 2.41				& 67.54				& 1.06				& N/A \\
		SVR				& 2.18				& 209.57			& 1.83				& 1.078 \\
		Gashler/Ashmore	& 2.81				& 94.47				& 3.71				& N/A \\
		ESN				& \textbf{1.09}		& 63.50				& 0.705				& N/A \\
		LSTM			& 1.14				& 93.61				& \textbf{0.667}	& N/A \\
		ND				& \textbf{1.09}		& \textbf{45.03}	& 0.99				& \textbf{0.214} \\ \hline
	\end{tabular}
	\label{table:rmse}
\end{table}

\section{Conclusion}
\label{sec:conclusion}

In this paper, we presented Neural Decomposition, a neural network technique for time-series forecasting.
Our method decomposes a set of training samples into a sum of sinusoids, inspired by the Fourier transform, augmented with additional components to enable our model to generalize and extrapolate beyond the input set.
Each component of the resulting signal is trained, so that it can find a simpler set of constituent signals.
ND uses careful initialization, input preprocessing, and regularization to facilitate the training process.
A toy problem was presented to demonstrate the necessity of each component of ND.
We applied ND to the Mackey-Glass series and was found to generalize well.
Finally, we showed results that demonstrate that our approach is superior to popular techniques LSTM, ESN, ARIMA, SARIMA, SVR, and Gashler and Ashmore's model in some cases, including the US unemployment rate, monthly airline passengers, and an unevenly sampled time-series of oxygen isotope measurements from a cave in north India.
We also showed that in some cases, our approach is at least comparible to these other techniques, as in the monthly time-series of ozone concentration in Los Angeles.
We predict that ND will similarly perform well on a number of other problems.

This work makes the following contributions to the current knowledge:

\begin{itemize}
	\item It empirically shows why the Fourier transform provides a poor initialization point for generalization and how neural network weights must be tuned to properly decompose a signal into its constituent parts.
	\item It demonstrates the necessity of an augmentation function in Fourier and Fourier-like neural networks and shows that components must be adjustable during the training process, observing the relationships between weight initialization, input preprocessing, and regularization in this context.
	\item It unifies these insights to describe a method for time-series forecasting and demonstrates that this method is effective at generalizing for some real-world datasets.
\end{itemize}

The primary area of future work is to apply ND to new problems.
The preliminary findings on the datasets in this paper show that ND can generalize well for some problems, but the breadth of applications for ND not yet known.
Some interesting areas to explore are traffic flow \cite{lippi:forecasting}, sales \cite{choi:sarima}, financial \cite{tay:financial}, and economic \cite{kaastra1996designing}.


\bibliographystyle{IEEEtran}
\bibliography{references}

\end{document}